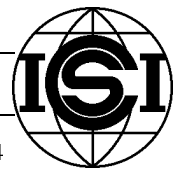

# Precise $n$-gram Probabilities from Stochastic Context-free Grammars

Andreas Stolcke*    Jonathan Segal†

TR-94-007

January 1994
(Revised April 1994)

**Abstract**

We present an algorithm for computing $n$-gram probabilities from stochastic context-free grammars, a procedure that can alleviate some of the standard problems associated with $n$-grams (estimation from sparse data, lack of linguistic structure, among others). The method operates via the computation of substring expectations, which in turn is accomplished by solving systems of linear equations derived from the grammar. We discuss efficient implementation of the algorithm and report our practical experience with it.

To appear in ACL-94.

*University of California at Berkeley, and International Computer Science Institute, 1947 Center Street, Berkeley, CA 94704, e-mail stolcke@icsi.berkeley.edu.
†Idem, e-mail jsegal@icsi.berkeley.edu.

# 1  Introduction

Probabilistic language modeling with $n$-gram grammars (particularly bigram and trigram) has proven extremely useful for such tasks as automated speech recognition, part-of-speech tagging, and word-sense disambiguation, and lead to simple, efficient algorithms. Unfortunately, working with these grammars can be problematic for several reasons: they have large numbers of parameters, so reliable estimation requires a very large training corpus and/or sophisticated smoothing techniques (Church & Gale 1991); it is very hard to directly model linguistic knowledge (and thus these grammars are practically incomprehensible to human inspection); and the models are not easily extensible, i.e., if a new word is added to the vocabulary, none of the information contained in an existing $n$-gram will tell anything about the $n$-grams containing the new item. Stochastic context-free grammars (SCFGs), on the other hand, are not as susceptible to these problems: they have many fewer parameters (so can be reasonably trained with smaller corpora); they capture linguistic generalizations, and are easily understood and written, by linguists; and they can be extended straightforwardly based on the underlying linguistic knowledge.

In this paper, we present a technique for computing an $n$-gram grammar from an existing SCFG—an attempt to get the best of both worlds. Besides developing the mathematics involved in the computation, we also discuss efficiency and implementation issues, and briefly report on our experience confirming its practical feasibility and utility.

The technique of compiling higher-level grammatical models into lower-level ones has precedents: Zue *et al.* (1991) report building a word-pair grammar from more elaborate language models to achieve good coverage, by random generation of sentences. In our own group, the current approach was predated by an alternative one that essentially relied on approximating bigram probabilities through Monte-Carlo sampling from SCFGs.

# 2  Preliminaries

An $n$-gram grammar is a set of probabilities $P(w_n|w_1w_2\ldots w_{n-1})$, giving the probability that $w_n$ follows a word string $w_1w_2\ldots w_{n-1}$, for each possible combination of the $w$'s in the vocabulary of the language. So for a 5000 word vocabulary, a bigram grammar would have approximately $5000 \times 5000 = 25,000,000$ free parameters, and a trigram grammar would have $\approx 125,000,000,000$. This is what we mean when we say $n$-gram grammars have many parameters.

A SCFG is a set of phrase-structure rules, annotated with probabilities of choosing a certain production given the left-hand side nonterminal. For example, if we have a simple CFG, we can augment it with the probabilities specified:

$$
\begin{array}{rcll}
S & \to & NP\ VP & [1.0] \\
NP & \to & N & [0.4] \\
NP & \to & Det\ N & [0.6] \\
VP & \to & V & [0.8] \\
VP & \to & V\ NP & [0.2] \\
Det & \to & \text{the} & [0.4] \\
Det & \to & \text{a} & [0.6] \\
N & \to & \text{book} & [1.0] \\
V & \to & \text{close} & [0.3] \\
V & \to & \text{open} & [0.7]
\end{array}
$$

The language this grammar generates contains 5 words. Including markers for sentence beginning and end, a bigram grammar would contain $6 \times 6$ probabilities, or $6 \times 5 = 30$ free parameters (since probabilities have to sum to one). A trigram grammar would come with $(5 \times 6 + 1) \times 5 = 155$ parameters. Yet, the above SCFG has only 10 probabilities, only 4 of which are free parameters.

The divergence between these two types of models generally grows as the vocabulary size increases, although this depends on the productions in the SCFG.

The reason for this discrepancy, of course, is that the *structure* of the SCFG itself is a discrete (hyper-)parameter with a lot of potential variation, but one that has been fixed beforehand. The point is that such a structure is comprehensible by humans, and can in many cases be constrained using prior knowledge, thereby reducing the estimation problem for the remaining probabilities. The problem of estimating SCFG parameters from data is solved with standard techniques, usually by way of likelihood maximization and a variant of the Baum-Welch (EM) algorithm (Baker 1979). A tutorial introduction to SCFGs and standard algorithms can be found in Jelinek *et al.* (1992).

## 3  Motivation

There are good arguments that SCFGs are in principle not adequate probabilistic models for natural languages, due to the conditional independence assumptions they embody (Magerman & Marcus 1991; Jones & Eisner 1992; Briscoe & Carroll 1993). Such shortcomings can be partly remedied by using SCFGs with very specific, semantically oriented categories and rules (Jurafsky *et al.* 1994). If the goal is to use $n$-grams nevertheless, then their their computation from a more constrained SCFG is still useful since the results can be interpolated with raw $n$-gram estimates for smoothing. An experiment illustrating this approach is reported later in the paper.

On the other hand, even if vastly more sophisticated language models give better results, $n$-grams will most likely still be important in applications such as speech recognition. The standard speech decoding technique of frame-synchronous dynamic programming (Ney 1984) is based on a first-order Markov assumption, which is satisfied by bigrams models (as well as by Hidden Markov Models), but not by more complex models incorporating non-local or higher-order constraints (including SCFGs). A standard approach is therefore to use simple language models to generate a preliminary set of candidate hypotheses. These hypotheses, e.g., represented as word lattices or $N$-best lists (Schwartz & Chow 1990), are re-evaluated later using additional criteria that can afford to be more costly due to the more constrained outcomes. In this type of setting, the techniques developed in this paper can be used to compile probabilistic knowledge present in the more elaborate language models into $n$-gram estimates that improve the quality of the hypotheses generated by the decoder.

Finally, comparing directly estimated, reliable $n$-grams with those compiled from other language models is a potentially useful method for evaluating the models in question.

For the purpose of this paper, then, we assume that computing $n$-grams from SCFGs is of either practical or theoretical interest and concentrate on the computational aspects of the problem.

It should be noted that there are alternative, unrelated methods for addressing the problem of the large parameter space in $n$-gram models. For example, Brown *et al.* (1992) describe an approach based on grouping words into classes, thereby reducing the number of conditional probabilities in the model.

## 4  The Algorithm

### 4.1  Normal form for SCFGs

A grammar is in *Chomsky Normal Form* (CNF) if every production is of the form $A \rightarrow B\ C$ or $A \rightarrow$ terminal. Any CFG or SCFG can be converted into one in CNF which generates exactly the same language, each of the sentences with exactly the same probability, and for which any parse in the original grammar would be reconstructible from a parse in the CNF grammar. In short, we can, without loss of generality, assume that the SCFGs we are dealing with are in CNF. In fact, our algorithm generalizes straightforwardly to the more general Canonical Two-Form (Graham *et al.* 1980) format, and in the case of bigrams ($n = 2$) it can even be modified to work directly for arbitrary SCFGs. Still, the CNF form is convenient, and to keep the exposition simple we assume all SCFGs to be in CNF.

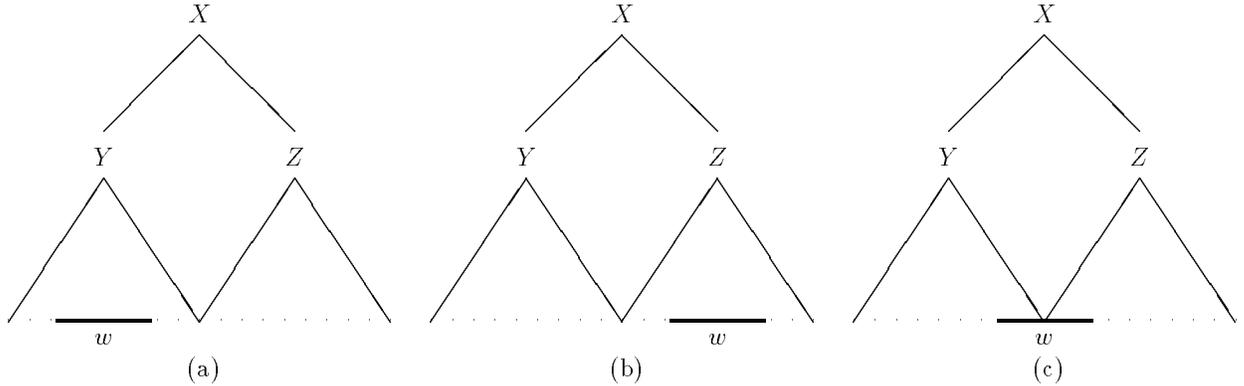

Figure 1: Three ways of generating a substring $w$ from a nonterminal $X$.

### 4.2 Probabilities from expectations

The first key insight towards a solution is that the $n$-gram probabilities can be obtained from the associated *expected frequencies* for $n$-grams and $(n-1)$-grams:

$$P(w_n|w_1w_2\ldots w_{n-1}) = \frac{c(w_1\ldots w_n|L)}{c(w_1\ldots w_{n-1}|L)} \quad (1)$$

where $c(w|L)$ stands for the expected count of occurrences of the substring $w$ in a sentence of $L$.[1]

*Proof.* Write the expectation for $n$-grams recursively in terms of those of order $n-1$ and the conditional $n$-gram probabilities:

$$c(w_1\ldots w_n|L) = c(w_1\ldots w_{n-1}|L)P(w_n|w_1w_2\ldots w_{n-1}).$$

So if we can compute $c(w|G)$ for all substrings $w$ of lengths $n$ and $n-1$ for a SCFG $G$, we immediately have an $n$-gram grammar for the language generated by $G$.

### 4.3 Computing expectations

Our goal now is to compute the substring expectations for a given grammar. Formalisms such as SCFGs that have a recursive rule structure suggest a divide-and-conquer algorithm that follows the recursive structure of the grammar.[2]

We generalize the problem by considering $c(w|X)$, the expected number of (possibly overlapping) occurrences of $w = w_1\ldots w_n$ in strings generated by an arbitrary nonterminal $X$. The special case $c(w|S) = c(w|L)$ is the solution sought, where $S$ is the start symbol for the grammar.

Now consider all possible ways that nonterminal $X$ can generate string $w = w_1\ldots w_n$ as a substring, denoted by $X \stackrel{*}{\Rightarrow} \ldots w_1\ldots w_n\ldots$, and the associated probabilities. For each production of $X$ we have to distinguish two main cases, assuming the grammar is in CNF. If the string in question is of length 1, $w = w_1$, *and* if $X$ happens to have a production $X \to w_1$, then that production adds exactly $P(X \to w_1)$ to the expectation $c(w|X)$.

---

[1] The only counts appearing in this paper are expectations, so be will not be using special notation to make a distinction between observed and expected values.

[2] A similar, even simpler approach applies to probabilistic finite state (i.e., Hidden Markov) models.

If $X$ has non-terminal productions, say, $X \to YZ$ then $w$ might also be generated by recursive expansion of the right-hand side. Here, for each production, there are three subcases.

(a) First, $Y$ can by itself generate the complete $w$ (see Figure 1(a)).

(b) Likewise, $Z$ itself can generate $w$ (Figure 1(b)).

(c) Finally, $Y$ could generate $w_1 \ldots w_j$ as a suffix ($Y \overset{*}{\Rightarrow}_R w_1 \ldots w_j$) and $Z$, $w_{j+1} \ldots w_n$ as a prefix ($Z \overset{*}{\Rightarrow}_L w_{j+1} \ldots w_n$), thereby resulting in a single occurrence of $w$ (Figure 1(c)).[3]

Each of these cases will have an expectation for generating $w_1 \ldots w_n$ as a substring, and the total expectation $c(w|X)$ will be the sum of these partial expectations. The total expectations for the first two cases (that of the substring being completely generated by $Y$ or $Z$) are given recursively: $c(w|Y)$ and $c(w|Y)$ respectively. The expectation for the third case is

$$\sum_{j=1}^{n-1} P(Y \overset{*}{\Rightarrow}_R w_1 \ldots w_j) P(Z \overset{*}{\Rightarrow}_L w_{j+1} \ldots w_n), \qquad (2)$$

where one has to sum over all possible split points $j$ of the string $w$.

To compute the total expectation $c(w|X)$, then, we have to sum over all these choices: the production used (weighted by the rule probabilities), and for each nonterminal rule the three cases above. This gives

$$\begin{aligned} c(w|X) &= P(X \to w) \\ &+ \sum_{X \to YZ} P(X \to YZ) \\ &\quad \bigg( c(w|Y) + c(w|Z) \\ &\quad + \sum_{j=1}^{n-1} P(Y \overset{*}{\Rightarrow}_R w_1 \ldots w_j) \\ &\quad\quad P(Z \overset{*}{\Rightarrow}_L w_{j+1} \ldots w_n) \bigg) \end{aligned} \qquad (3)$$

In the important special case of bigrams, this summation simplifies quite a bit, since the terminal productions are ruled out and splitting into prefix and suffix allows but one possibility:

$$\begin{aligned} c(w_1 w_2 | X) &= \sum_{X \to YZ} P(X \to YZ) \\ &\quad \bigg( c(w_1 w_2 | Y) + c(w_1 w_2 | Z) \\ &\quad + P(Y \overset{*}{\Rightarrow}_R w_1) P(Z \overset{*}{\Rightarrow}_L w_2) \bigg) \end{aligned} \qquad (4)$$

For unigrams equation (3) simplifies even more:

$$\begin{aligned} c(w_1 | X) &= P(X \to w_1) \\ &+ \sum_{X \to YZ} P(X \to YZ) \bigg( c(w_1|Y) + c(w_1|Z) \bigg) \end{aligned} \qquad (5)$$

---

[3] We use the notation $X \overset{*}{\Rightarrow}_R \alpha$ to denote that non-terminal $X$ generates the string $\alpha$ as a suffix, and $X \overset{*}{\Rightarrow}_L \alpha$ to denote that $X$ generates $\alpha$ as a prefix. Thus $P(X \overset{*}{\Rightarrow}_L \alpha)$ and $P(X \overset{*}{\Rightarrow}_R \alpha)$ are the probabilities associated with those events.

We now have a recursive specification of the quantities $c(w|X)$ we need to compute. Alas, the recursion does not necessarily bottom out, since the $c(w|Y)$ and $c(w|Z)$ quantities on the right side of equation (3) may depend themselves on $c(w|X)$. Fortunately, the recurrence is linear, so for each string $w$, we can find the solution by solving the linear system formed by all equations of type (3). Notice there are exactly as many equations as variables, equal to the number of nonterminals in the grammar. The solution of these systems is further discussed below.

## 4.4 Computing prefix and suffix probabilities

The only substantial problem left at this point is the computation of the constants in equation (3). These are derived from the rule probabilities $P(X \to w)$ and $P(X \to YZ)$, as well as the prefix/suffix generation probabilities $P(Y \stackrel{*}{\Rightarrow}_R w_1 \ldots w_j)$ and $P(Z \stackrel{*}{\Rightarrow}_L w_{j+1} \ldots w_n)$.

The computation of prefix probabilities for SCFGs is generally useful for applications, and has been solved with the LRI algorithm (Jelinek & Lafferty 1991). Recently, Stolcke (1993) has shown how to perform this computation efficiently for sparsely parameterized SCFGs using a probabilistic version of Earley's parser (Earley 1970). Computing suffix probabilities is obviously a symmetrical task; for example, one could create a 'mirrored' SCFG (reversing the order of right-hand side symbols in all productions) and then run any prefix probability computation on that mirror grammar.

Note that in the case of bigrams, only a particularly simple form of prefix/suffix probabilities are required, namely, the 'left-corner' and 'right-corner' probabilities, $P(X \stackrel{*}{\Rightarrow}_L w_1)$ and $P(Y \stackrel{*}{\Rightarrow}_R w_2)$, which can each be obtained from a single matrix inversion (Jelinek & Lafferty 1991).

It should be mentioned that there are some technical conditions that have to be met for a SCFG to be well-defined and consistent (Booth & Thompson 1973). These condition are also sufficient to guarantee that the linear equations given by (3) have positive probabilities as solutions. The details of this are discussed in the Appendix.

Finally, it is interesting to compare the relative ease with which one can solve the substring *expectation* problem to the seemingly similar problem of finding substring *probabilities*: the probability that $X$ generates (one or more instances of) $w$. The latter problem is studied by Corazza *et al.* (1991), and shown to lead to a *non-linear* system of equations. The crucial difference here is that expectations are additive with respect to the cases in Figure 1, whereas the corresponding probabilities are not, since the three cases can occur simultaneously.

## 5 Efficiency and Complexity Issues

Summarizing from the previous section, we can compute any $n$-gram probability by solving two linear systems of equations of the form (3), one with $w$ being the $n$-gram itself and one for the $(n-1)$-gram prefix $w_1 \ldots w_{n-1}$. The latter computation can be shared among all $n$-grams with the same prefix, so that essentially one system needs to be solved for each $n$-gram we are interested in. The good news here is that the work required is linear in the number of $n$-grams, and correspondingly limited if one needs probabilities for only a subset of the possible $n$-grams. For example, one could compute these probabilities on demand and cache the results.

Let us examine these systems of equations one more time. Each can be written in matrix notation in the form

$$(\mathbf{I} - \mathbf{A})\mathbf{c} = \mathbf{b} \qquad (6)$$

where $\mathbf{I}$ is the identity matrix, $\mathbf{A} = (a_{XU})$ is a coefficient matrix, $\mathbf{b} = (b_X)$ is the right-hand side vector, and $\mathbf{c}$ represents the vector of unknowns, $c(w|X)$. All of these are indexed by nonterminals $X, U$.

We get

$$a_{XU} = \sum_{X \to YZ} P(X \to YZ)(\delta(Y,U) + \delta(Z,U)) \qquad (7)$$

$$\begin{aligned}
b_X &= P(X \to w) \\
&+ \sum_{X \to YZ} P(X \to YZ) \\
&\qquad \sum_{j=1}^{n-1} P(Y \stackrel{*}{\Rightarrow}_R w_1 \ldots w_j) \\
&\qquad\qquad P(Z \stackrel{*}{\Rightarrow}_L w_{j+1} \ldots w_n) \qquad (8)
\end{aligned}$$

where $\delta(X, Y) = 1$ if $X = Y$, and 0 otherwise. The expression $\mathbf{I} - \mathbf{A}$ arises from bringing the variables $c(w|Y)$ and $c(w|Z)$ to the other side in equation (3) in order to collect the coefficients.

We can see that all dependencies on the particular bigram, $w$, are in the right-hand side vector **b**, while the coefficient matrix $\mathbf{I} - \mathbf{A}$ depends only on the grammar. This, together with the standard method of *LU decomposition* (see, e.g., Press *et al.* (1988)) enables us to solve for each bigram in time $O(N^2)$, rather than the standard $O(N^3)$ for a full system ($N$ being the number of nonterminals/variables). The LU decomposition itself is cubic, but is incurred only once. The full computation is therefore dominated by the quadratic effort of solving the system for each $n$-gram. Furthermore, the quadratic cost is a worst-case figure that would be incurred only if the grammar contained every possible rule; empirically we have found this computation to be linear in the number of nonterminals, for grammars that are *sparse*, i.e., where each nonterminal makes reference only to a bounded number of other nonterminals.

## 6  Summary

Listed below are the steps of the complete computation. For concreteness we give the version specific to bigrams ($n = 2$).

1. Compute the prefix (left-corner) and suffix (right-corner) probabilities for each (nonterminal,word) pair.

2. Compute the coefficient matrix and right-hand sides for the systems of linear equations, as per equations (4) and (5).

3. LU decompose the coefficient matrix.

4. Compute the unigram expectations for each word in the grammar, by solving the LU system for the unigram right-hand sides computed in step 2.

5. Compute the bigram expectations for each word pair by solving the LU system for the bigram right-hand sides computed in step 2.

6. Compute each bigram probability $P(w_2|w_1)$, by dividing the bigram expectation $c(w_1w_2|S)$ by the unigram expectation $c(w_1|S)$.

## 7  Experiments

The algorithm described here has been implemented, and is being used to generate bigrams for a speech recognizer that is part of the BeRP spoken-language system (Jurafsky *et al.* 1994). An early prototype of BeRP was used in an experiment to assess the benefit of using bigram probabilities obtained through SCFGs versus estimating them directly from the available training corpus.[4] The system's domain are inquiries about restaurants in the city of Berkeley. The training corpus used had only 2500 sentences, with an average length of about 4.8 words/sentence.

---

[4] Corpus and grammar sizes, as well as the recognition performance figures reported here are not up-to-date with respect to the latest version of BeRP. For ACL-94 we expect to have revised results available that reflect the current performance of the system.

Our experiments made use of a context-free grammar hand-written for the BeRP domain. With 1200 rules and a vocabulary of 1100 words, this grammar was able to parse 60% of the training corpus. Computing the bigram probabilities from this SCFG takes about 24 hours on a SPARCstation 2-class machine.[5]

In experiment 1, the recognizer used bigrams that were estimated directly from the training corpus, without any smoothing, resulting in a word error rate of 35.1%. In experiment 2, a different set of bigram probabilities was used, computed from the context-free grammar, whose probabilities had previously been estimated from the same training corpus, using standard EM techniques. This resulted in a word error rate of 35.3%. This may seem surprisingly good given the low coverage of the underlying CFGs, but notice that the conversion into bigrams is bound to result in a less constraining language model, effectively increasing coverage.

Finally, in experiment 3, the bigrams generated from the SCFG were augmented by those from the raw training data, in a proportion of 200,000 : 2500. We have not attempted to optimize this mixture proportion, e.g., by deleted interpolation (Jelinek & Mercer 1980).[6] With the bigram estimates thus obtained, the word error rate dropped to 33.5%. (All error rates were measured on a separate test corpus.)

The experiment therefore supports our earlier argument that more sophisticated language models, even if far from perfect, can improve $n$-gram estimates obtained directly from sample data.

## 8 Conclusions

We have described an algorithm to compute in closed form the distribution of $n$-grams for a probabilistic language given by a stochastic context-free grammar. Our method is based on computing substring expectations, which can be expressed as systems of linear equations derived from the grammar. The algorithm was used successfully and found to be practical in dealing with context-free grammars and bigram models for a medium-scale speech recognition task, where it helped to improve bigram estimates obtained from relatively small amounts of data.

Deriving $n$-gram probabilities from more sophisticated language models appears to be a generally useful technique which can both improve upon direct estimation of $n$-grams, and allow available higher-level linguistic knowledge to be effectively integrated into the speech decoding task.

## Acknowledgments

Dan Jurafsky wrote the BeRP grammar, carried out the recognition experiments, and was generally indispensable. Steve Omohundro planted the seed for our $n$-gram algorithm during lunch at the California Dream Café by suggesting substring expectations as an interesting computational linguistics problem. Thanks also to Jerry Feldman and Lokendra Shastri for improving the presentation with their comments.

This research was partially supported by ARPA contract #N0000 1493 C0249.

---

[5] Unlike the rest of BeRP, this computation is implemented in Lisp/CLOS and could be speeded up considerably if necessary.

[6] This proportion comes about because in the original system, predating the method described in this paper, bigrams had to be estimated from the SCFG by random sampling. Generating 200,000 sentence samples was found to give good converging estimates for the bigrams. The bigrams from the raw training sentences were then simply added to the randomly generated ones. We later verified that the bigrams estimated from the SCFG were indeed identical to the ones computed directly using the method described here.

# A  Appendix: Consistency of SCFGs

Blindly applying the $n$-gram algorithm (and many others) to a SCFG with arbitrary probabilities can lead to surprising results. Consider the following simple grammar

$$\begin{aligned} S &\rightarrow x & [p] \\ S &\rightarrow SS & [q = 1-p] \end{aligned} \qquad (9)$$

What is the expected frequency of unigram $x$? Using the abbreviation $c = c(X|S)$ and equation 5, we see that

$$\begin{aligned} c &= P(S \rightarrow x) + P(S \rightarrow SS)(c + c) \\ &= p + 2qc \end{aligned}$$

This leads to

$$c = \frac{p}{1-2q} = \frac{p}{2p-1}. \qquad (10)$$

Now, for $p = 0.5$ this becomes infinity, and for probabilities $p < 0.5$, the solution is negative! This is a rather striking manifestation of the failure of this grammar, for $p \leq 0.5$, to be *consistent*. A grammar is said to be inconsistent if the underlying stochastic derivation process has non-zero probability of not terminating (Booth & Thompson 1973). The expected length of the generated strings should therefore be infinite in this case.

Fortunately, Booth and Thompson derive a criterion for checking the consistency of a SCFG: Find the first-order expectancy matrix $\mathbf{E} = (e_{XY})$, where $e_{XY}$ is the expected number of occurrences of nonterminal $Y$ in a one-step expansion of nonterminal $X$, and make sure its powers $\mathbf{E}^k$ converge to 0 as $k \rightarrow \infty$. If so, the grammar is consistent, otherwise it is not.[7]

For the grammar in (9), $\mathbf{E}$ is the $1 \times 1$ matrix $(2q)$. Thus we can confirm our earlier observation by noting that $(2q)^k$ converges to 0 iff $q < 0.5$, or $p > 0.5$.

Now, it so happens that $\mathbf{E}$ is identical to the matrix $\mathbf{A}$ that occurs in the linear equations (6) for the $n$-gram computation. The actual coefficient matrix is $\mathbf{I} - \mathbf{A}$, and its inverse, if it exists, can be written as the geometric sum

$$(\mathbf{I} - \mathbf{A})^{-1} = \mathbf{I} + \mathbf{A} + \mathbf{A}^2 + \mathbf{A}^3 + \ldots$$

This series converges precisely if $\mathbf{A}^k$ converges to 0. We have thus shown that the existence of a solution for the $n$-gram problem is equivalent to the consistency of the grammar in question. Furthermore, the solution vector $\mathbf{c} = (\mathbf{I} - \mathbf{A})^{-1}\mathbf{b}$ will always consist of non-negative numbers: it is the sum and product of the non-negative values given by equations (7) and (8).

---

[7] A further version of this criterion is to check the magnitude of the largest of $\mathbf{E}$'s eigenvalues (its spectral radius). If that value is $> 1$, the grammar is inconsistent; if $< 1$, it is consistent.